\documentclass[letterpaper]{article} % DO NOT CHANGE THIS
\usepackage{aaai2026}  % DO NOT CHANGE THIS
\usepackage{times}  % DO NOT CHANGE THIS
\usepackage{helvet}  % DO NOT CHANGE THIS
\usepackage{courier}  % DO NOT CHANGE THIS
\usepackage[hyphens]{url}  % DO NOT CHANGE THIS
\usepackage{graphicx} % DO NOT CHANGE THIS
\usepackage{amsfonts}
\usepackage{multirow}

\usepackage{natbib}  % DO NOT CHANGE THIS AND DO NOT ADD ANY OPTIONS TO IT
\usepackage{caption} % DO NOT CHANGE THIS AND DO NOT ADD ANY OPTIONS TO IT
\usepackage{algorithm}
\usepackage{algorithmic}
\usepackage{booktabs}
\usepackage{amsmath}  % 确保导言区引入了它

\usepackage{newfloat}
\usepackage{listings}
\DeclareCaptionStyle{ruled}{labelfont=normalfont,labelsep=colon,strut=off} % DO NOT CHANGE THIS
\floatstyle{ruled}
\newfloat{listing}{tb}{lst}{}
\floatname{listing}{Listing}
\title{Multi-Treatment-DML: Causal Estimation for Multi-Dimensional Continuous Treatments with Monotonicity Constraints in Personal Loan Risk Optimization}
\author{
    Kexin Zhao\textsuperscript{\rm 1}, 
    Bo Wang\textsuperscript{\rm 1}, 
    Cuiying Zhao\textsuperscript{\rm 1},
    Tongyao Wan\textsuperscript{\rm 1}
}
\affiliations{
    \textsuperscript{\rm 1}Didi Chuxing\\
    zhaokexinlexi@gmail.com, \{edenwang, zhaocuiying, wangtongyao\}@didiglobal.com
}

\usepackage{bibentry}
\begin{document}

\maketitle

\begin{abstract}
Optimizing credit limits, interest rates, and loan terms is crucial for managing borrower risk and lifetime value (LTV) in personal loan platform. 
However, counterfactual estimation of these continuous, multi-dimensional treatments faces significant challenges: 
randomized trials are often prohibited by risk controls and long repayment cycles, forcing reliance on biased observational data. 
Existing causal methods primarily handle binary/discrete treatments and struggle with continuous, multi-dimensional settings. 
Furthermore, financial domain knowledge mandates provably monotonic treatment-outcome relationships (e.g., risk increases with credit limit).
To address these gaps, we propose Multi-Treatment-DML, a novel framework leveraging Double Machine Learning (DML) to: 
(i) debias observational data for causal effect estimation; 
(ii) handle arbitrary-dimensional continuous treatments; 
and (iii) enforce monotonic constraints between treatments and outcomes, guaranteeing adherence to domain requirements.
Extensive experiments on public benchmarks and real-world industrial datasets demonstrate the effectiveness of our approach. 
Furthermore, online A/B testing conducted on a realworld personal loan platform, 
confirms the practical superiority of Multi-Treatment-DML in real-world loan operations.
\end{abstract}


\section{Introduction}

Individual-level causal inference (ICI) focuses on estimating causal effects at the unit level by quantifying potential outcomes under alternative treatments, 
enabling the estimation of Individual Treatment Effects (ITE~\cite{yao2021survey}). 
ICI has significant applications across domains including healthcare, marketing, and finance. 
Within personal lending, accurately estimating counterfactual risk under varying credit limits, interest rates, 
and loan terms is critical for optimizing risk management and customer Lifetime Value (LTV). 
This work addresses the challenging problem of continuous, multi-dimensional treatment effect estimation from observational data in this high-stakes domain.

While extensive research exists on causal effect estimation, 
most methods (e.g., meta-learners) assume access to abundant Randomized Controlled Trial (RCT) data. 
In personal lending, however, RCTs are often infeasible due to high financial risk and prolonged outcome observation periods. 
Consequently, we must rely solely on observational data, which introduces significant challenges due to confounding bias. 

A primary confounder in this domain is borrower creditworthiness: 
platforms systematically assign treatments-including credit limits, rates, terms-based on unobserved or partially observed applicant characteristics, 
inducing selection bias in the observed data.
The systematic correlation between unobserved creditworthiness factors and treatment assignment creates endogeneity 
that confounds the estimation of true causal effects,
thereby inflating observed treatment effects and leading to spurious causal conclusions. 
This selection mechanism necessitates the development of robust methodologies that can 
disentangle the causal impact of financial interventions from the confounding influence of borrower characteristics.

Core methodologies for addressing confounding in observational data aim to balance feature distributions across treatment groups: 
Inverse Probability Weighting (IPW)~\cite{rosenbaum1983ipw} reweights samples to achieve balance, 
while Double Machine Learning (DML)~\cite{chernozhukov2017doubleml} employs residualization to orthogonalize treatment and outcome variables against confounders. 
Representation learning approaches (e.g., DeR-CFR~\cite{wu2020dercfr}) seek balanced representations in latent space. However, significant gaps persist:

\begin{itemize}
  \item \textbf{Treatment Complexity:} Most methods target binary or discrete treatments. VCNet handles continuous treatments but assumes RCT settings, 
  limiting applicability to observational financial data.
  \item \textbf{Multi-Dimensionality:} Methods for multi-dimensional continuous treatments remain underdeveloped.
  \item \textbf{Domain Constraints:} Financial risk modeling requires strict adherence to domain-specific monotonicity constraints
  (e.g., higher credit limits must correlate with higher risk), which existing methods rarely enforce.
\end{itemize}

To bridge these gaps, we propose Multi-Treatment DML, a novel framework extending Double Machine Learning. Our key contributions are:
\begin{itemize}
  \item \textbf{Generalized Orthogonalization for Multi-Dimensional Treatments:} 
  We extend the DML paradigm through sequential residualization, simultaneously orthogonalizing multiple continuous treatment variables against confounders. 
  This explicitly eliminates confounding bias while accommodating arbitrary treatment dimensions inherent to credit policies (limit, rate, term).
  \item \textbf{Monotonicity-Constrained Uplift Estimation:} We introduce a monotonicity module within the outcome network, 
  enforcing provable monotonic relationships between treatments (e.g., credit limit) and outcomes (e.g., default risk). 
  This guarantees model adherence to critical financial domain knowledge, enhancing reliability and trustworthiness.
  \item \textbf{Rigorous Empirical Validation:} We demonstrate Multi-Treatment DML's efficacy through extensive offline experiments on public benchmarks and a large-scale proprietary lending dataset. Furthermore, we validate its practical impact via large-scale online experiments on a fintech platform serving over 2 million users, confirming significant improvements in risk estimation and business metrics.
\end{itemize}

\section{Related Works}

Uplift modeling has a wide range of application domains, 
including precision marketing~\cite{liu2023explicit, mukerji2024valuing}, 
medical interventions~\cite{ma2023look}, and personalized product~\cite{guelman2015decision}. 

In practice, uplift modeling can be implemented through various approaches, 
the type of available data, the complexity of the treatment variables, the requirements for model interpretability, and the actual needs.
Classical approaches, such as S-Learner, T-Learner, and X-Learner~\cite{kunzel2019metalearners}, 
decompose the estimation of  effects into separate or joint models for treated and control groups. 
These meta-learning frameworks typically rely on RCT data to ensure unconfoundedness. 
However, in financial domains, the implementation of RCTs is often infeasible due to ethical, regulatory, 
or business constraints, thereby limiting the practical applicability of these methods.

To overcome the scarcity of RCT data, recent research has focused on causal inference from observational datasets, 
where treatment assignment is inherently subject to selection bias. 
Double Machine Learning (DML)~\cite{chernozhukov2017doubleml} employs orthogonalization and sample splitting to control for confounding, 
while representation learning methods such as TARNet and DragonNet~\cite{shalit2017tarnet, shi2019dragonnet} aim to learn balanced feature representations that mitigate bias. 
In recent years, representation learning methods such as D2VD~\cite{kuang2017D2VD}, DR-CFR~\cite{wu2020dercfr}, TEDVAE~\cite{zhang2021tedvae},
and MIN-DRCFR~\cite{cheng2022MIMDRCFR} have been proposed to learn disentangled latent factors. 
These methods aims to disentangle representations into components that capture variables affecting only the treatment (instrumental variables), only the outcome (adjustment variables), 
and those influencing both the treatment and the outcome (confounders).
% In the context of personal lending, instrumental variables ($I$) are those that influence the treatment assignment $T$ 
% but do not directly affect the outcome $Y$ except through $T$, such as regional policy changes. 
% Confounders ($C$) are variables that affect both the treatment assignment and the outcome, 
% introducing bias if not properly controlled; typical examples include credit score, income level, and repayment history, 
% Adjustment variables ($A$) are predictive of the outcome $Y$ but are independent of the treatment $T$, 
% such as stable behavioral features.
Nevertheless, existing methods predominantly 
focus on binary treatment scenarios and are not directly applicable to continuous treatment problems.

To address the challenge of continuous treatments, researchers have proposed several methods. 
DRNet~\cite{schwab2020drnet} and VCNet~\cite{nie2021vcnet} extend neural network architectures to accommodate continuous treatments.
DRNet~\cite{schwab2020drnet} addresses continuous treatment effect estimation by discretizing the treatment variable into $E$ intervals, 
with each branch of the network learning the outcome for a specific treatment interval.
DRNet~\cite{schwab2020drnet} addresses continuous treatment effect estimation by discretizing the treatment variable into $E$ intervals, 
with each branch of the network learning the outcome for a specific treatment interval.
VCNet~\cite{nie2021vcnet} is an improvement based on DRNet, 
adopting a varying coefficient model that allows the coefficients of the outcome model to flexibly change as a function of the continuous treatment. 
This design enables direct and interpretable uplift estimation for continuous treatments, 
capturing nuanced dose-response relationships in observational data.
However, these designs are primarily tailored for RCT settings and do not directly address confounding bias present in observational data.

Another critical challenge is ensuring that model outputs satisfy monotonicity constraints required by business logic. 
In our case, it is essential for model predictions to strictly adhere to monotonicity requirements.
This reflects fundamental business principles and enhances both model interpretability and reliability in practical deployment. 
Several approaches have been proposed, such as Unconstrained Monotonic Neural Networks (UMNN)~\cite{wehenkel2019umnn}, 
which are based on the insight that a function is monotonic as long as its derivative is strictly positive. 
Lipschitz Monotonic Networks (LMN)~\cite{kitouni2023expressive} use non-negative weights 
and a residual connection to guarantee monotonicity with respect to selected input features. 
Another approach, Constrained Monotone Fully Connected Layers~\cite{runje2023cmnn}, 
achieves monotonicity by designing special activation functions.
However, these methods are often complex and rely on deep neural network architectures, 
operate as black-box models, which limits their interpretability. 
In high-stakes financial applications, interpretability is crucial for regulatory compliance, risk management, and business decision-making. 
Therefore, it is important to design monotonicity modules that not only enforce domain constraints 
but also provide transparent and explainable model behavior.

% In summary, existing methods are limited in financial personal lending scenarios. 
% Most approaches are designed for RCT data or only support binary/discrete treatments, 
% while RCTs are often infeasible in finance due to cost, long observation periods, and business risk. 
% Real-world financial interventions are typically continuous and multi-dimensional, 
% which conventional methods do not handle well. 
% Additionally, current models rarely enforce monotonicity constraints or provide interpretable outputs, 
% both of which are essential for financial risk modeling.

\section{Preliminary Propositions}

We follow the potential outcomes framework for causal inference proposed by Rubin~\cite{rubin2005causal}, 
to define the problem of estimating risk of in a multi-dimensional continuous treatment setting.
Considering an observational dataset in a financial context, 
denoted as $\mathcal{D} = \{(\mathbf{x}_i, \mathbf{t}_i, \mathit{y}_i)\}_{i=1}^N$ 
sampled from a joint distribution $p(\textbf{X}, \textbf{T}, \mathit{Y})$
where $Y = f(\textbf{X}, \textbf{T}) + \varepsilon$

\begin{figure}[ht]
  \centering
  \includegraphics[width=0.8\linewidth]{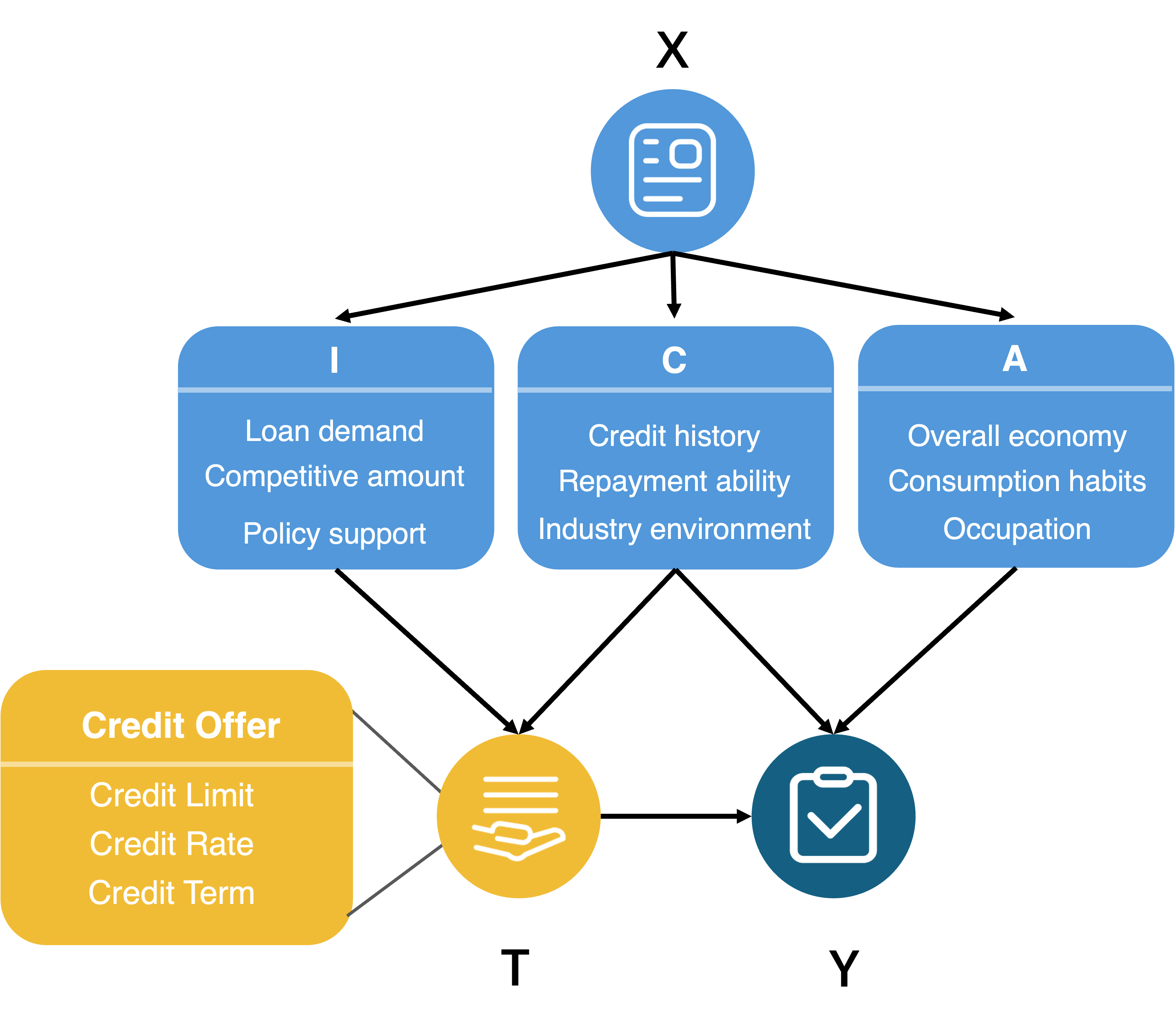}
  \caption{Illustration of the causal inference framework for multi-dimensional continuous treatments in Personal Loan modeling.}
  \label{fig:causal_framework}
\end{figure}

As illustrated in Figure~\ref{fig:causal_framework}, 
$\textbf{X} \in \mathbb{R}^D$ denotes the observed covariates—user characteristics and behavioral features available prior to intervention. 
In the context of personal loan, \textbf{X} includes fundamental attributes and historical behaviors, 
such as actual cost, overdue counts, total repayments, and other relevant user information.
The treatment vector $\textbf{T} \in \mathbb{R}^{K}$ represents $K$-dimensional continuous financial offer variables 
(e.g., credit limit, interest rate, loan term with $K=3$), 
which directly affect the outcome $\textit{Y} \in \mathbb{R} $. 
Unlike binary or categorical interventions ($K=1$, $\textit{T} \in \{0,1\}$), 
financial treatments are inherently multi-dimensional and continuous. 
The outcome variable $\textit{Y}$ denotes the observed response,
such as utilized loan amount.
It is important to emphasize that our dataset consists solely of observational data, 
where treatment assignments are determined by internal business rules rather than randomized protocols.

The estimation of individual treatment effects (ITE) relies on several key assumptions~\cite{rosenbaum1983ipw}:

\begin{itemize}
  \item \textbf{Assumption 1 (SUTVA):} \textit{Stable Unit Treatment Value Assumption.} 
  The outcome for any individual depends only on their own treatment assignment 
  and is unaffected by the treatment assignments of others.
  \item \textbf{Assumption 2 (Unconfoundedness):} \textit{Conditional Independence.} Given observed covariates $\textbf{X}$, 
  the treatment assignment is independent of the potential outcomes: $(\textit{Y}_0, \textit{Y}_1) \perp\!\!\!\perp T \mid \textbf{X}$.
  \item \textbf{Assumption 3 (Positivity):} \textit{Overlap.} 
  Every individual has a non-zero probability of receiving each possible treatment given their covariates: 
  $0 < P(\textit{T} = t \mid \textbf{X} = x) < 1$ for all $t$ and $x$.
\end{itemize}

\section{Methodology}

\begin{figure*}[ht]
  \centering
  \includegraphics[width=0.62\linewidth]{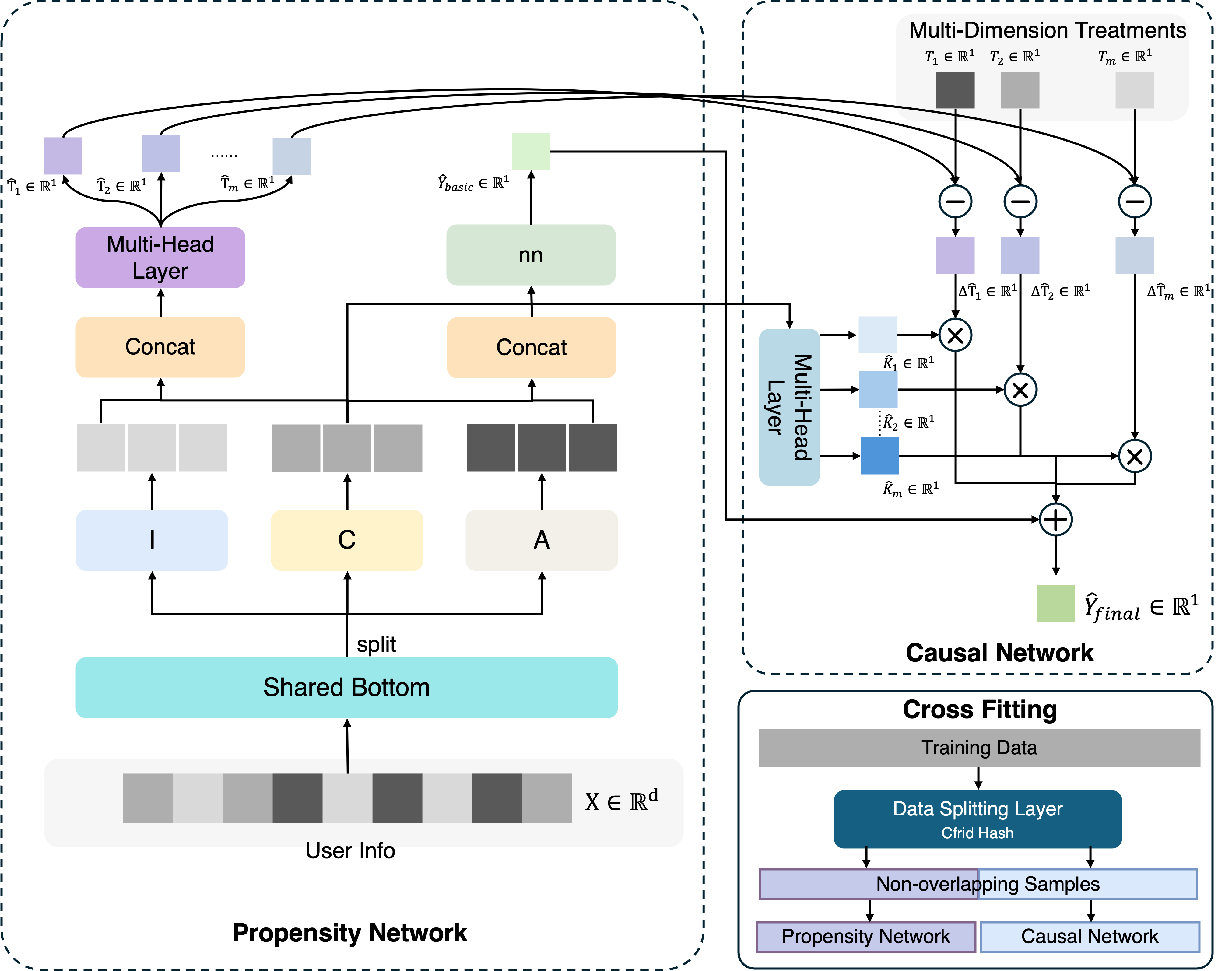}
  \caption{Illustration of the overall architecture of Multi-Treatment DML for multi-dimensional continuous treatments.}
  \label{fig:overall_architecture}
\end{figure*}

As shown in the Figure~\ref{fig:overall_architecture}, our proposed Multi-Treatment DML framework consists of two main components: the \textbf{Propensity Network} and the \textbf{Causal Network}. The Propensity Network is responsible for modeling the treatment assignment mechanism and disentangling latent factors from observed features, while the Causal Network estimates the treatment effect and enforces monotonicity constraints. In the following subsections, we provide a detailed description of each component.

\subsection{Propensity Network}

The Propensity Network predicts the treatment propensity $\hat{\mathbf{T}} = \mathbf{E}[\mathbf{T}|\mathbf{X}]$, 
the outcome $\hat{\textit{Y}} = \mathbf{E}[\textit{Y}|\textbf{X}]$, 
and a sensitivity coefficient $K$—which is used in the causal network to enforce monotonicity between $\Delta \textbf{T}$ and $\Delta \textit{Y}$—
from the observed covariates $X$. 

For generality, we assume that the feature set $\mathbf{X}$ in the dataset comprises all relevant observed variables, 
which can be partitioned into instrumental variables ($\textbf{I}$), confounders ($\textbf{C}$), and adjustment variables ($\textbf{A}$), 
i.e., $\{$\textbf{I}$, $\textbf{C}$, $\textbf{A}$\}$. 
We aim to encode user features $\textbf{X} \in \mathbb{R}^D$ into three independent latent embeddings 
corresponding to instrumental variables ($\textbf{I}$), confounders ($\textbf{C}$), and adjustment variables ($\textbf{A}$), 
denoted as $\textbf{I}, \textbf{C}, \textbf{A} = \mathrm{Disentangle}(\textbf{X})$.
To achieve this, we employ a shared-bottom neural architecture that first extracts a common feature representation from $\textbf{X}$, 
which is then decoded by three factor-specific layers into the respective latent factors. 
The disentanglement is guided by following tasks: 

\textbf{\textit{Treatment Prediction: }}
Let $\boldsymbol{\omega} = \mathrm{Concat}(\mathbf{I}, \mathbf{C})$ denote the concatenation of instrumental and confounder representations. 
We use a treatment prediction network $\mathcal{F}_{\mathbf{T}}$ to predict treatment,
minimizing the prediction error between the observed treatment $\mathbf{T}$ and the predicted $\hat{\mathbf{T}}$.
For multi-dimensional treatments ($\mathbf{T} \in \mathbb{R}^{K}$), the loss generalizes to:
\begin{equation}
\mathcal{L}_{\text{treatment}} = \frac{1}{N} \sum_{i=1}^N \sum_{k=1}^K \ell(\mathrm{T}_{i,k}, \mathrm{\hat{T}}_{i,k})
\end{equation}

\textbf{\textit{Outcome Prediction: }}
Similarly, let $\boldsymbol{\Phi} = \mathrm{Concat}(\mathbf{C}, \mathbf{A})$ denote the 
concatenation of confounder and adjustment representations. 
We use an outcome prediction network $\mathcal{F}_{\textit{Y}}$ to Predict the outcome $\textit{Y}$.
The training objective for the outcome prediction is to 
minimize the prediction error between the observed outcome $\textit{Y}$ and the predicted $\hat{Y}$:
\begin{equation}
\mathcal{L}_{\text{outcome}} = \frac{1}{N} \sum_{i=1}^N \ell(Y_i, \hat{Y}_i)
\end{equation}

\begin{figure}[ht]
  \centering
  \includegraphics[width=1.0\linewidth]{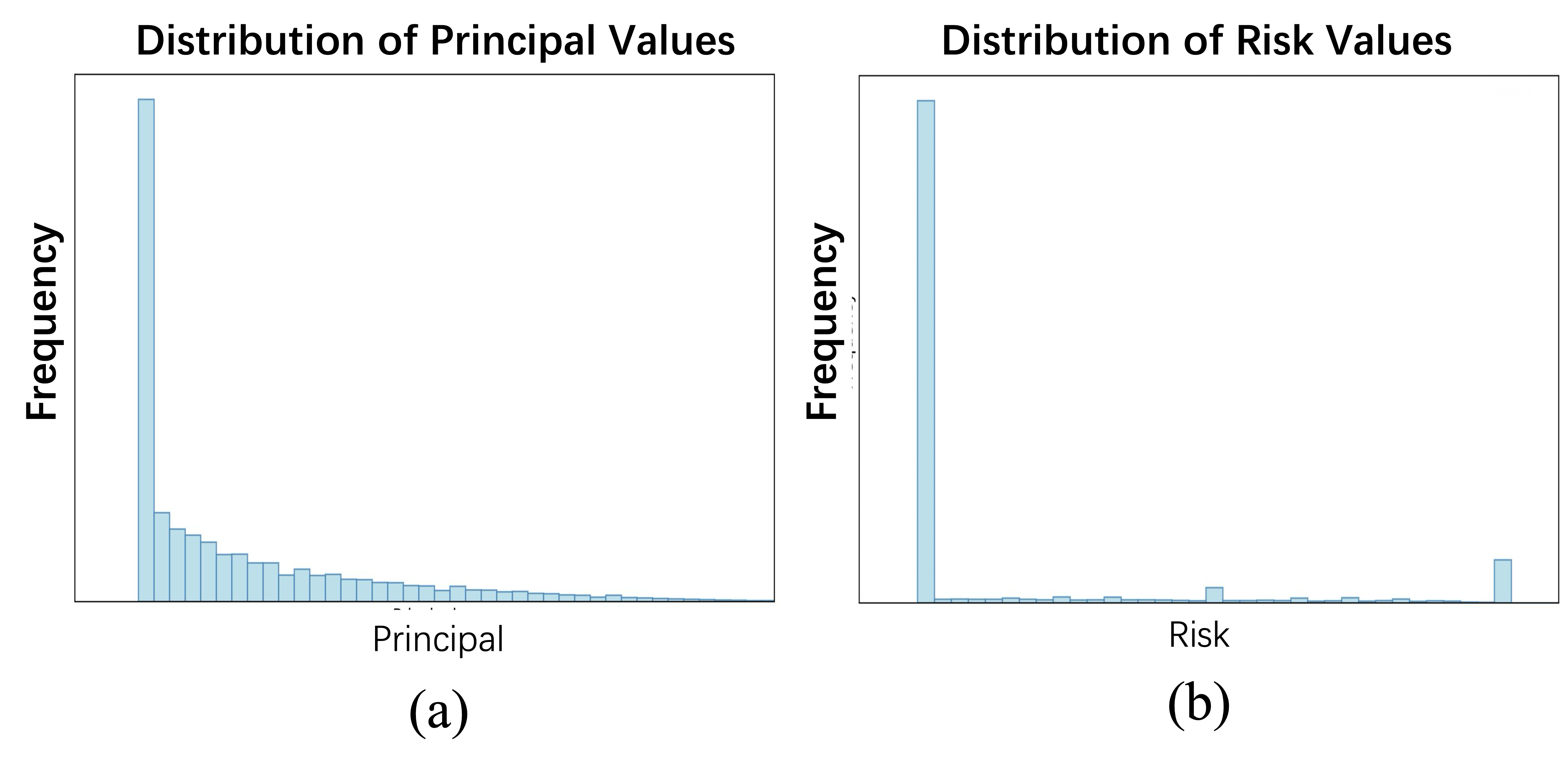}
  \caption{Distribution of outcome variable in the Personal Loan dataset: (a) Principle Value, (b) Risk Value.}
  \label{fig:long_tail_distribution}
\end{figure}
Figure~\ref{fig:long_tail_distribution} illustrates the distribution of 
both Principle Value and Risk Value in the Personal Loan dataset. 
Both variables exhibit pronounced long-tail characteristics:
most samples are concentrated at lower values with very few at higher values, 
and zero-inflation, with a substantial proportion of samples accumulating at zero, reflecting users with no transaction or utilization. 
These properties should be considered when designing outcome prediction objectives and model architectures,
as standard loss functions may not adequately capture the distributional nuances of such variables.
Therefore, we employ the Tweedie loss function for outcome prediction, which is well-suited for modeling long-tailed, zero-inflated financial variables. The Tweedie loss is defined as:

\begin{equation}
\text{Tweedie} = \frac{1}{N} \sum_{i=1}^N \left( Y_i \frac{\hat{Y}_i^{1-\rho}}{1-\rho} - \frac{\hat{Y}_i^{2-\rho}}{2-\rho} \right)
\end{equation}
In our experiments, we set the Tweedie power parameter $\rho$ to $1.9$,
which is commonly used for modeling zero-inflated, long-tailed financial outcomes. 
\textbf{\textit{Weight Orthogonality Regularization: }}
We introduce a criterion called Representation Layer Orthogonality (RLO), 
which provides an intuitive approach to obtaining non-overlapping factors:
\begin{equation}
\small
\mathcal{L}_{\text{RLO}} = \cos(\bar{\mathrm{W}}_{I}, \bar{\mathrm{W}}_{C}) + \cos(\bar{\mathrm{W}}_{C}, \bar{\mathrm{W}}_{A}) + \cos(\bar{\mathrm{W}}_{A}, \bar{\mathrm{W}}_{I})
\end{equation}
where $\cos(\cdot, \cdot)$ denotes the cosine similarity between two weight vectors. 
Minimizing $\mathcal{L}_{\text{RLO}}$ encourages the learned representations 
for $\mathbf{I}$, $\mathbf{C}$, and $\mathbf{A}$ to be mutually orthogonal, 
thus promoting disentanglement and ensuring that each latent factor 
captures distinct information from the input features.

\subsection{Causal Network}
The Causal Network is designed to model and learn the monotonic relationship 
between $\mathbf{\Delta T}$ and $\mathbf{\Delta Y}$. 
Enforcing this monotonicity ensures model interpretability and business applicability, 
and effectively prevents counter-intuitive or unreasonable prediction results.

\textbf{\textit{Sensitivity Coefficient $K$ Prediction: }}
In addition to treatment and outcome prediction, 
we introduce a sensitivity coefficient $K$ to quantify the monotonic relationship 
between treatment changes and outcome risk,
which is critical for enforcing monotonicity in the subsequent Causal Network. 
we predict the sensitivity coefficient 
$K$ using a sensitivity prediction network $\mathcal{F}_K$ from $\Upsilon = \mathrm{Concat}(\mathbf{I}, \mathbf{C}, \mathbf{A})$

To prevent the sensitivity coefficient $K$ from becoming excessively large 
and to promote model stability, we introduce an $L_2$ regularization term on $K$:
\begin{equation}
\mathcal{L}_{K\text{-reg}} = \frac{1}{N} \sum_{i=1}^N \|K_i\|^2
\end{equation}
This term constrains the magnitude of $K$, 
ensuring that the monotonic effect of treatment on outcome remains 
reasonable and avoids multicollinearity, which can lead to the phenomenon of squeezing between factors.

\textbf{\textit{Monotonicity between $\mathbf{\Delta T}$ and $\mathbf{\Delta Y}$ :}} 
Given the stringent monotonicity requirements inherent in our business context, 
we formalize this relationship using a linear parameterization. 
and has the abilit to ensures reliable and interpretable results in practice.
Specifically, we model the change in predicted outcome $\mathbf{\Delta Y}$ 
as a monotonic function of the change in treatment $\mathbf{\Delta T}$, 
parameterized by a sensitivity coefficient $K$ that is dynamically predicted by the Propensity Network. 
This formulation ensures that any increase in a treatment component $\mathbf{T_k}$ 
cannot result in a decrease in the predicted outcome $Y$, thereby rigorously enforcing monotonicity by design.
Given the predicted outcome $Y$, the linear monotonic model $f_{\text{linear}}(K, \Delta T)$ 
outputs the final predicted outcome $\hat{Y}_{\text{final}}$:
$\hat{Y}_{\text{final}} = f_{\text{linear}}(K, \mathbf{\Delta T}) + Y$

The loss function can be defined as the mean squared error between the true outcome $Y$ and the sum $\hat{Y}_{\text{final}}$:
\begin{equation}
\mathcal{L}_{\text{final}} = \frac{1}{N} \sum_{i=1}^N \ell(Y_i, \hat{Y}_{\text{final},i})
\end{equation}
Here, we also use Tweedie loss for outcome prediction.

\subsection{Overall Loss Function}
The overall loss function for Multi-Treatment DML combines the objectives from both the Propensity Network and the Causal Network, 
along with regularization terms. Formally, the total loss is defined as:

\begin{align}
\mathcal{L}_{\text{total}} =\ & \mathcal{L}_{\text{treatment}} + \mathcal{L}_{\text{outcome}} + \mathcal{L}_{\text{final}} \nonumber \\
& + \lambda_{\text{RLO}} \mathcal{L}_{\text{RLO}} + \lambda_{K} \mathcal{L}_{K\text{-reg}}
\end{align}

\subsection{Cross-Fitting}

$K$-fold cross-fitting  is a crucial step in DML to mitigate estimation bias caused by overfitting. 
In the original DML procedure, the dataset is split into n disjoint subsets,
Next, select the next sample to estimate residuals, and repeat the above steps until all n samples are processed. 
Finally, average the two estimates to obtain the final result. 
However, implementing $K$-fold cross-fitting in neural network-based Multi-Treatment DML is non-trivial 
due to the complexity of model training and parameter sharing across folds. 
the data is split into two non-overlapping parts: one part is used to train the Propensity Network, 
and the other to train the Causal Network. 
The process is repeated with the folds swapped, 
and the final estimate is averaged over both rounds. 

% \begin{algorithm}[ht]
% \caption{Cross-Fitting Procedure for Multi-Treatment DML}
% \begin
% {algorithmic}[1]
% \REQUIRE Dataset $\mathcal{D} = \{(x_i, t_i, y_i)\}_{i=1}^N$
% \STATE Split $\mathcal{D}$ into 2 folds: $\mathcal{D}_1, \mathcal{D}_2$
% \FOR{both fold $k = 1$ to $k = 2$}
%   \STATE Let $\mathcal{D}_{1}$ be all data except $\mathcal{D}_k$
%   \STATE Train Propensity Network on $\mathcal{D}_{-k}$ 
%   \STATE Train Causal Network on $\mathcal{D}_k$
% \ENDFOR
% \STATE Aggregate treatment effect estimates from 2 folds 
% \RETURN Final treatment effect estimate
% \end{algorithmic}
% \end{algorithm}

\section{Experiments}
In this section, we present a comprehensive evaluation of the proposed Multi-Treatment DML framework. 
We first introduce the datasets and evaluation metrics used in our experiments. 
Then, we report comparative results on public benchmark datasets, followed by results on a real-world financial lending dataset, 
including an ablation study to analyze the contribution of each module.

\subsection{Datasets}
To evaluate the performance and effectiveness of the proposed Multi-Treatment DML method,
we conduct experiments on three public datasets—Twins~\cite{almond2005costs},
JOBS~\cite{lalonde1986evaluating}, and Lazada~\cite{zhong2022descn}
Additionally, we include a large-scale financial dataset from a personal loan platform to demonstrate real-world applicability.

% \textbf{Twins:} The Twins dataset is a semi-synthetic dataset constructed from US twin births, 
% including 46 pre-intervention variables and the one-year mortality outcome. It is commonly used for benchmarking causal inference algorithms.

% \textbf{JOBS:} The JOBS dataset originates from a randomized field experiment on the effect of a job training program, 
% containing treatment assignments and employment outcomes.

% \textbf{Lazada Dataset:} We further validate our model on a real-world e-commerce dataset from Lazada,
% where the treatment is coupon distribution. The dataset includes both observational and RCT-based subsets, 
% allowing for robust evaluation under selection bias.

% \textbf{Personal Loan Dataset:} 
% We utilize a proprietary dataset collected from a large-scale online personal loan platform. 
% Each data record corresponds to a loan order and includes detailed user demographic and behavioral features 
% (e.g., age, income, repayment history), 
% multi-dimensional continuous treatment variables—specifically credit limit, interest rate, and loan term
% —as well as observed repayment outcomes such as utilized amount and default probability. 
% The data is sliced at the order level with an observation window of three months per record.

% Treatments are assigned according to internal business rules rather than randomized protocols, 
% leading to selection bias and non-random treatment assignment. 
% All sensitive information has been thoroughly anonymized 
% and processed in strict compliance with applicable data privacy regulations

\subsection{Evaluation Metrics}
For datasets that contain ground truth individual treatment effect $\tau_t(x)$, 
we adopt the expexted Precision in Estimation of Heterogeneous Effect ($\epsilon_{\mathrm{PEHE}}$) 
as the evaluation metric, 
as well as the Absolute Error of Average Treatment Effect ($\epsilon_{\mathrm{ATE}}$).
% formatlly defined as follows:
% \begin{equation}
%   \epsilon_{\mathrm{PEHE}} = \sqrt{ \frac{1}{n} \sum_{i=1}^n \left( \left( \hat{\mu}_1(x_i) - \hat{\mu}_0(x_i) \right) - \tau(x_i) \right)^2 }
% \end{equation}

% \begin{equation}
%   \epsilon_{\mathrm{ATE}} = \frac{1}{n} \sum_{i=1}^n \left( \hat{\mu}_1(x_i) - \hat{\mu}_0(x_i) \right) - \frac{1}{n} \sum_{i=1}^n \tau(x_i)
% \end{equation}
$\epsilon_{\mathrm{PEHE}}$ is a widely used metric to evaluate the accuracy of treatment effect estimation at the individual level,
Meanwhile, $\epsilon_{\mathrm{ATE}}$ serves as a more suitable metric for evaluating the difference in average treatment effects between sample groups.

For datasets where ground truth uplift values are not available,
we evaluate model performance using Qini AUUC,
the Error of Average Treatment effect on the Treated ($\epsilon_{\mathrm{ATT}}$), and policy risk. 
% AUUC is computed by plotting the uplift curve as described above and calculating the area under this curve, 
% normalized by the total number of samples. 
% This metric reflects the model's ability to rank individuals according to their predicted treatment effect.
% Because all the treated subjects $T$ were part of the original randomized sample $E$, 
% we can compute the true average treatment effect on the treated (ATT) as follows, 
% where $C$ is the control group:
% \begin{equation}
%   \mathrm{ATT} = \frac{1}{|T|} \sum_{i \in T} y_i - \frac{1}{|C \cap E|} \sum_{i \in C \cap E} y_i
% \end{equation}

% We reported the error of average treatment effect on the treated ($\epsilon_{\mathrm{ATT}}$) as follows:
% \begin{equation}
%   \epsilon_{\mathrm{ATT}} = \left| \frac{1}{|T|} \sum_{i \in T} \left( \hat{\mu}_1(x_i) - \hat{\mu}_0(x_i) \right) - \mathrm{ATT} \right|
% \end{equation}

% For a given model $f$, we define the treatment policy $\pi_f(x)$ as follows: 
% assign treatment ($\pi_f(x) = 1$) if the predicted uplift $f(x, 1) - f(x, 0) > \lambda$, 
% otherwise do not assign treatment ($\pi_f(x) = 0$). In this study, we set $\lambda = 0$.
% The policy risk is defined as follows:

% \begin{equation}
% \small
% \begin{aligned}
%     & \mathbb{E}[Y_1 \mid \pi_f(x) = 1] \cdot p(\pi_f = 1) \\
%     & + \mathbb{E}[Y_0 \mid \pi_f(x) = 0] \cdot p(\pi_f = 0)
% \Big)
% \end{aligned}
% \end{equation}

\subsection{Comparative Results on Publication Datasets}

In Table~\ref{tab:twins_results}, 
Multi-Treatment-DML achieves the best individual treatment effect estimation accuracy 
($\sqrt{\epsilon_{\text{PEHE}}}$) and average treatment effect error ($\epsilon_{\text{ATE}}$) 
on the Twins dataset, with values of $0.323 \pm 0.002$ and $0.011 \pm 0.001$, respectively, 
significantly outperforming all baseline methods. 

Table~\ref{tab:jobs_lazada_results} presents a comparative evaluation of Multi-Treatment DML 
and baseline methods on the Jobs and Lazada datasets. 
For the Jobs dataset, Multi-Treatment DML achieves the lowest policy risk ($\mathcal{R}_{\mathrm{pol}}(\pi) = 0.221$) 
and ATT error ($\epsilon_{\text{ATT}} = 0.064$), outperforming all baselines. 
On the Lazada dataset, Multi-Treatment DML matches the best policy risk ($0.993$) 
and achieves competitive ATT error ($0.043$), while Qini AUUC ($0.018$) is comparable to the best baseline. 
Overall, Multi-Treatment-DML demonstrates superior or competitive results across all metrics, 
confirming its effectiveness in both policy optimization and treatment effect estimation.
More importantly, our method is capable of handling arbitrary-dimensional 
and continuous treatments setting where existing methods are rarely applicable.
This represents a significant advantage and the key innovation of our approach.

% \begin{table}[h]
% \centering
% \renewcommand{\arraystretch}{1.2} 
% \begin{tabular}{lcc}
% \toprule
% \multirow{2}{*}{\textbf{Model}} & $\sqrt{\epsilon_{\text{PEHE}}}$ & $\epsilon_{\text{ATE}}$ \\
% & mean $\pm$ s.e. & mean $\pm$ s.e. \\
% \midrule 
% T-Learner    & $0.319 \pm 0.003$ & $0.015 \pm 0.002$ \\
% S-Learner    & $0.321 \pm 0.003$ & $0.006 \pm 0.002$ \\
% X-Learner    & $0.322 \pm 0.003$ & $0.009 \pm 0.002$ \\
% DR-Learner   & $0.337 \pm 0.001$ & $0.004 \pm 0.001$ \\
% TNet         & $0.341 \pm 0.005$ & $0.053 \pm 0.012$ \\
% SNet         & $0.340 \pm 0.002$ & $0.047 \pm 0.009$ \\
% XNet         & $0.338 \pm 0.003$ & $0.056 \pm 0.012$ \\
% RNet         & $0.339 \pm 0.003$ & $0.052 \pm 0.009$ \\
% \midrule 
% TARNet       & $0.336 \pm 0.002$ & $0.033 \pm 0.009$ \\
% DragonNet    & $0.338 On the \textbf{Jobs dataset}, Multi-Treatment-DML achieves the best policy risk (${\mathcal{R}_{\mathrm{pol}}(\pi)} = 0.221 \pm 0.009$) 
% EFIN         & $0.338 \pm 0.005$ & $0.026 \pm 0.004$ \\
% GANITE*      & $0.297 \pm 0.160$   & $-$ \\
% DeR-CFR*     & $0.279 \pm 0.005$  & $0.008 \pm 0.004$ \\
% \midrule
% \textbf{MTDML}  & $0.323 \pm 0.002$ & $0.011 \pm 0.001$ \\
% \bottomrule
% \end{tabular} 
% \caption{Performance comparison on the Twins dataset.  
% (The star marks means the results reported in the original paper that used the same replications.)}
% \label{tab:epilepsy_results_simplified}
% \end{table}

\begin{table}[h]
\centering
\renewcommand{\arraystretch}{1.2} 
\begin{tabular}{lcc}
\toprule
\multirow{2}{*}{\textbf{Model}} & $\sqrt{\epsilon_{\text{PEHE}}}$ & $\epsilon_{\text{ATE}}$ \\
& mean $\pm$ s.e. & mean $\pm$ s.e. \\
\midrule 
T-Learner         & $0.341 \pm 0.005$ & $0.053 \pm 0.012$ \\
S-Learner          & $0.340 \pm 0.002$ & $0.047 \pm 0.009$ \\
X-Learner          & $0.338 \pm 0.003$ & $0.056 \pm 0.012$ \\
\midrule 
TARNet       & $0.336 \pm 0.002$ & $0.033 \pm 0.009$ \\
DragonNet    & $0.338 \pm 0.003$ & $0.035 \pm 0.009$ \\
EFIN          & $0.338 \pm 0.005$ & $0.026 \pm 0.004$ \\
\midrule
\textbf{MTDML}  & $\mathbf{0.323 \pm 0.002}$ & $\mathbf{0.011 \pm 0.001}$ \\
\bottomrule
\end{tabular} 
\caption{Performance comparison on the Twins dataset.  
(The star marks mean the results reported in the original paper that used the same replications.)}
\label{tab:twins_results}
\end{table}

\begin{table*}[h]
\centering
\renewcommand{\arraystretch}{1.2}
\begin{tabular}{lcc|ccc}
\toprule
\multirow{3}{*}{\textbf{Model}} & \multicolumn{2}{c|}{\textbf{Jobs Dataset}} & \multicolumn{3}{c}{\textbf{Lazada Dataset}} \\
& ${\mathcal{R}_{\mathrm{pol}}(\pi)}$ & $\epsilon_{\text{ATT}}$
& ${\mathcal{R}_{\mathrm{pol}}(\pi)}$ & $\epsilon_{\text{ATT}}$ & {\textbf{Qini AUUC}} \\
& mean $\pm$ s.e. & mean $\pm$ s.e. 
& mean $\pm$ s.e. & mean $\pm$ s.e. & mean \\
\midrule
T-Learner   & $0.389 \pm 0.060$ & $0.105 \pm 0.034$ & $0.992 \pm 0.000$ & $0.042 \pm 0.000$ & $0.002 \pm 0.000$ \\
S-Learner   & $0.345 \pm 0.068$ & $0.117 \pm 0.041$ & $\mathbf{0.990 \pm 0.000} $ & $0.042 \pm 0.000$ & $\mathbf{0.099 \pm 0.000}$ \\
X-Learner   & $0.565 \pm 0.110$ & $0.084 \pm 0.037$ & $0.985 \pm 0.000$ & $0.043 \pm 0.000$ & $0.054 \pm 0.001$\\
\midrule
TARNet      & $0.479 \pm 0.108$ & $0.180 \pm 0.052$ & $0.987 \pm 0.001$ & $0.075 \pm 0.006$ & $0.002 \pm 0.000$ \\
DragonNet   & $0.768 \pm 0.098$ & $0.192 \pm 0.031$ & $0.987 \pm 0.002$ & $0.071 \pm 0.009$ & $0.018 \pm 0.027$\\
EFIN        & $0.294 \pm 0.045$ & $0.107 \pm 0.038$ & $0.993 \pm 0.000$ & $\mathbf{0.017 \pm 0.002}$ & $0.002 \pm 0.000$ \\
\midrule
\textbf{MTDML}  & $\mathbf{0.221 \pm 0.009}$ & $\mathbf{0.064 \pm 0.024}$ & $\mathbf{0.993 \pm 0.000}$  & $0.043 \pm 0.001$ & $0.018 \pm 0.001$ \\
\bottomrule
\end{tabular}
\caption{Performance comparison on the Jobs and Lazada datasets.}
\label{tab:jobs_lazada_results}
\end{table*}

\subsection{Overall Performance on Personal Loan Dataset}

\subsubsection{Visualization of $\Delta T$ and  $\Delta Y$}

\begin{figure}[ht]
  \centering
  \includegraphics[width=0.9\linewidth]{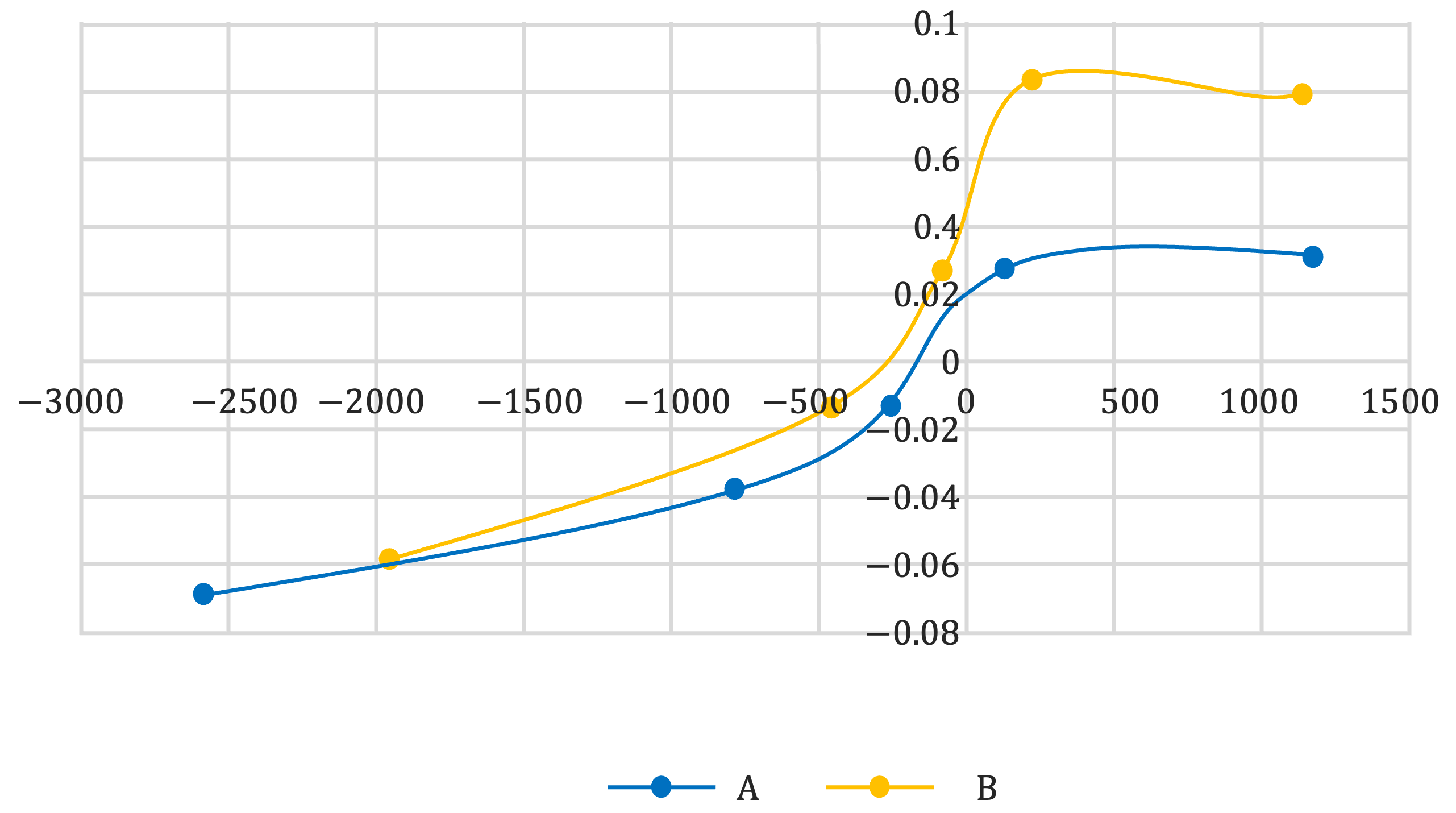}
  \caption{Trend of $\Delta Y$ versus $\Delta T$ for the outcome variable. User group etc.}

  \label{fig:deltaT_trend}
\end{figure}

Figure~\ref{fig:deltaT_trend} visualizes the risk variation trends for two distinct creditworthiness groups 
(high-credit group A and medium-credit group B) under different credit limit increases. 
Consistent with financial risk domain expertise, 
the sensitivity of risk to credit limit increases is significantly lower 
for high-credit customers compared to medium-credit customers. 
The results demonstrate that as $\Delta T$ (credit limit increase) grows, $\Delta Y$ (risk change) 
exhibits a linear upward trend, with the slope for group B being notably steeper. 
This finding empirically validates the business-side hypothesis. 
Moreover, the observed linear relationship between $\Delta T$ and $\Delta Y$ indicates 
that the proposed framework effectively eliminates selection bias arising from user creditworthiness, 
enabling accurate modeling of heterogeneous risk sensitivities across customer segments
 and further substantiating the framework's debiasing capability and practical applicability.

\subsubsection{Ablation Study}

We conduct an ablation study on the real-world online business dataset to evaluate the contribution of each module in our framework. 
The purpose of the ablation study is to systematically assess the impact of key components:

\begin{itemize}
  \item \textbf{DML}: The original Double Machine Learning implementation using the public \texttt{econml} library.
  \item \textbf{DML-NN}: DML implemented with neural network architectures for both propensity and outcome models.
  \item \textbf{DML-NN-ICA}: DML-NN with additional disentanglement to partition latent factors 
  into instrumental variables ($\textbf{I}$), confounders ($\textbf{C}$), and adjustment variables ($\textbf{A}$).
  \item \textbf{DML-NN-ICA-Tweedie}: DML-NN-ICA further enhanced by using Tweedie loss as the prediction objective for the outcome function.
\end{itemize}

For evaluation, we adopt the PCOC (Predict Click Over Click) metric.
PCOC (Predict Click Over Click) is a commonly used metric in online advertising to evaluate the calibration of predicted click-through rates (CTR).
It is defined as the ratio between the average predicted CTR and the average observed (empirical) CTR:
A PCOC value close to 1 indicates well-calibrated predictions,
and values less than 1 indicate underestimation.
As shown in Table~\ref{tab:uplift_ablation}, the best results, with PCOC closest to $1$ ($1.002$) 
and the highest Qini AUUC ($0.187$). 
These results indicate that the proposed enhancements lead to more accurate and better-calibrated uplift modeling, 
with Qini AUUC increasing and PCOC approaching $1$, which is optimal.

\begin{table}[h]
\centering
\begin{tabular}{lcc}
\toprule
\textbf{Model} & \textbf{PCOC} & \textbf{Qini AUUC} \\
\midrule
DML   & $1.07$ & $0.103$ \\
DML-NN   & $1.11$ & $0.165$ \\
DML-NN-ICA   & $1.09$ & $0.168$ \\
DML-NN-ICA-Tweedie   & $\textbf{1.002}$ & $\textbf{0.187}$ \\
\bottomrule
\end{tabular}
\caption{Ablation study results on the online business dataset.}
\label{tab:uplift_ablation}
\end{table}

\subsubsection{Comparative Results}
In this experiment, we simplify the problem setting for all baseline models by 
treating the increase in credit limit versus no increase as a binary treatment. 
This allows for a fair comparison between our proposed Multi-Treatment DML 
and other uplift modeling approaches under a unified scenario.

Table~\ref{tab:personal_loan_results} presents the comparative results of various methods on the Personal Loan dataset in terms of PCOC and Qini AUUC metrics. 
As shown, Multi-Treatment-DML achieves the highest Qini AUUC score of $0.243$, 
significantly outperforming all baseline methods, indicating its superior uplift modeling and ranking performance. Meanwhile, Multi-Treatment-DML's PCOC is $1.100$, 
which is closer to the ideal calibration value of $1$, demonstrating better consistency between predicted and actual click-through rates. 
Other methods, such as DragonNet and TarNet, also perform well on Qini AUUC, but their PCOC values deviate more from $1$, 
suggesting some degree of underestimation. Overall, Multi-Treatment-DML balances prediction accuracy and ranking capability in the personal loan scenario, 
validating the effectiveness and business value of the proposed method.
\begin{table}[h]
\centering
\begin{tabular}{lcc}
\toprule
\textbf{Model} & \textbf{PCOC} & \textbf{Qini AUUC} \\
\midrule
S-Learner   & $0.825$ & $0.214$ \\
T-Learner   & $0.845$ & $0.122$ \\
X-Learner   & $0.830$ & $0.156$ \\
\midrule
TARNet   & $0.891$ & $0.203$ \\
DragonNet   & $0.885$ & $0.197$ \\
EFIN   & $0.826$ & $0.060$ \\
MTDML   & $\textbf{1.100}$ & $\textbf{0.243}$ \\
\bottomrule
\end{tabular}
\caption{Comparative results on the Personal Loan dataset.}
\label{tab:personal_loan_results}
\end{table}

\begin{figure}[ht]
  \centering
  \includegraphics[width=1.0\linewidth]{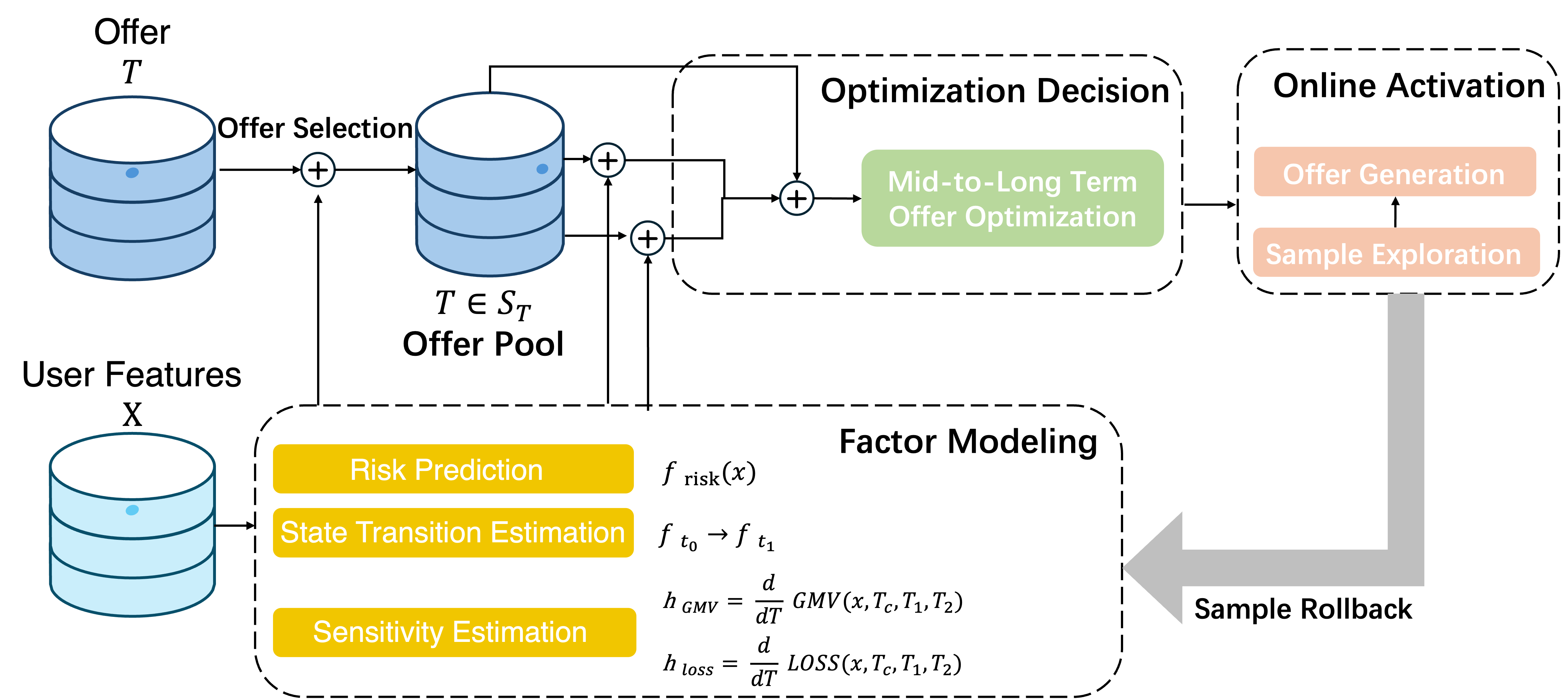}
  \caption{Overview of the online deployment framework for Multi-Treatment DML in personal loan risk optimization.}
  \label{fig:Online Deployment}
\end{figure}

To validate the real-world business value of our proposed method, 
we deployed it in a production online credit decision system, 
aiming to optimize long-term business returns through refined strategy selection. 
As illustrated in Figure~\ref{fig:Online Deployment}, the overall deployment framework consists of three main stages:

\begin{itemize}
  \item \textbf{Candidate Offer Generation:} The system generates a set of credit offers 
  (including credit limit, interest rate, and repayment term) for each user based on their features.
  \item \textbf{Factor Modeling:} Leveraging the multi-Treatment-DML architecture for heterogeneous factor modeling (e.g., GMV and principal loss), 
  our mechanism-based computational model generates counterfactual LTV estimations  of each candidate offer. 
  And select the best one to maximize mid-to-long term LTV.
  \item \textbf{Mid-to-Long Term LTV Optimization (Three Months):} The system selects and delivers the offer for users 
  to maximizes mid-to-long term value (specifically, over a three-month horizon).
\end{itemize}

\section{Online Deployment Performance}

While executing the optimal strategy, the system reserves a certain proportion for exploration 
to enable continuous learning of the strategy space. Subsequent user behaviors are fed back for model iteration, 
forming a closed-loop optimization process.In practice, we applied this method to multiple versions of online strategy tests. 
Results show that the LTV-guided strategy achieved a 10\% cumulative return improvement over the baseline, 
with the advantage further increasing over longer observation periods, 
demonstrating the effectiveness of our approach in optimizing long-term business value in real-world settings.

Table~\ref{tab:ltv_comparison} summarizes the results of the online deployment 
comparing the baseline strategy with the Multi-Treatment DML approach. 
The Multi-Treatment DML strategy achieved a cumulative profit gain of 10\% over the baseline, 
demonstrating a significant improvement in business performance. 
This uplift indicates that the Multi-Treatment DML framework 
effectively optimizes credit strategies, 
leading to higher profitability compared to traditional approaches. T
he consistent profit gain across multiple online experiments validates the practical value
and robustness of the proposed method in real-world financial applications.
\begin{table}[h]
\centering
\caption{Comparison of Cumulative Profit Gains Across Strategy Versions}
\begin{tabular}{lc}
\toprule
\textbf{Strategy Version} & \textbf{Cumulative Profit Gain} \\
\midrule
Baseline  & Baseline \\
Multi—TreatmentDML& \textbf{+10\%} \\
\bottomrule
\end{tabular}
\label{tab:ltv_comparison}
\end{table}

\section{Conclusion}
In this paper, we proposed Multi-Treatment-DML, 
a novel framework for causal estimation under multi-dimensional continuous treatments with monotonicity constraints, 
specifically tailored for personal loan risk optimization. 
Our approach extends Double Machine Learning to handle complex, 
high-dimensional interventions and incorporates monotonic neural modules to enforce domain-specific constraints, 
ensuring interpretability and reliability in financial applications. 

Extensive experiments on public benchmarks and a large-scale real-world lending dataset demonstrate that 
Multi-Treatment-DML achieves superior performance compared to existing methods. 
Our results demonstrate that Multi-Treatment-DML achieves SOTA
or near-SOTA performance across multiple evaluation metrics. 
More importantly, the primary contribution and novelty of our work lie 
in the ability of our framework to handle both multi-dimensional and continuous-valued treatments—a setting 
for which existing methods are rarely applicable,
marking a substantial advancement in causal inference research.
Furthermore, successful online deployment validates its practical value, yielding significant improvements in business metrics.
Future work will explore broader applications and further enhancements to address more complex causal structures 
in financial decision-making.

\bibliography{Multi_Treatment_DML}

\end{document}